\documentclass{article}
\usepackage{ijcai21}

\usepackage{times}
\usepackage{soul}
\usepackage{url}
\usepackage[hidelinks]{hyperref}
\usepackage[utf8]{inputenc}
\usepackage[small]{caption}
\usepackage{graphicx}
\usepackage{amsmath}
\usepackage{amsthm}
\usepackage{booktabs}
\usepackage{algorithm}
\usepackage{algorithmic}
\usepackage{xspace}
\usepackage{graphics}
\usepackage{adjustbox}

\newcommand{\oov}{\textsc{oov}\xspace}
\newcommand{\finsim}{\textsc{finsim}\xspace}
\newcommand{\finbert}{\textsc{finbert}\xspace}

\newcommand{\dicoe}{\textsc{DICoE}\xspace}

\title{DICoE@FinSim-3: Financial Hypernym Detection using Augmented Terms and Distance-based Features}

\author{
Lefteris Loukas$^{1,2}$ \and
Konstantinos Bougiatiotis$^1$\and
Manos Fergadiotis$^{1,2}$ \and
Dimitris Mavroeidis$^1$ \and
Elias Zavitsanos$^1$
\\
\affiliations
$^1$Institute of Informatics and Telecommunications, \\National Centre for Scientific Research ``Demokritos", Greece \\
$^2$ Department of Informatics, Athens University of Economics and Business, Greece
\emails
\{eloukas, bogas.ko, mfergadiotis, dmavroeidis, izavits\}@iit.demokritos.gr}

\begin{document}

\maketitle

\begin{abstract}
We present the submission of team \dicoe for \finsim-3, the 3rd Shared Task on Learning Semantic Similarities for the Financial Domain. The task provides a set of terms in the financial domain and requires to classify them into the most relevant hypernym from a financial ontology. After augmenting the terms with their Investopedia definitions, our system employs a Logistic Regression classifier over financial word embeddings and a mix of hand-crafted and distance-based features.
Also, for the first time in this task, we employ different replacement methods for out-of-vocabulary terms, leading to improved performance.
Finally, we have also experimented with word representations generated from various financial corpora. 
Our best-performing submission ranked 4th on the task's leaderboard.
\end{abstract}

\section{Introduction}
Taxonomies constitute the backbone of many knowledge representation schemas, especially in the context of the Semantic web and ontologies. Such hierarchies model among others the \textit{hypernymy} relation, a significant semantic relation between concepts. It is an asymmetric relation between two concepts, a hyponym (subordinate) and a hypernym (superordinate), as in ``car-vehicle'', where the hyponym necessarily implies the hypernym, but not vice versa.

\begin{figure}[h]
  \centering
  \includegraphics[width=\linewidth]{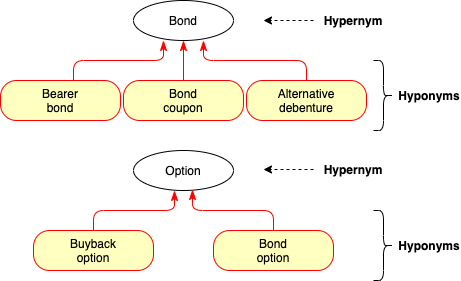}
  \caption{Examples of hypernym relations from the FIBO ontology. Interestingly, ``Bond coupon" is a kind of ``Bond", but ``Bond option" is a kind of ``Option".}
  \label{fig:hypernymy}
\end{figure}

In the context of hypernym detection, the shared task on Learning Semantic Similarities for the Financial Domain (\finsim-3) focuses on the evaluation of semantic representations by assessing the classification of a given list of terms from the financial domain against a domain ontology. A list of carefully selected terms from the financial domain is provided, such as ``European depositary receipt", ``Interest rate swaps", and others, and the task is to design a system that can classify them into the most relevant hypernym (or top-level) concept in an external ontology. The referenced ontology is the Financial Industry Business Ontology (FIBO).\footnote{Visit \url{https://spec.edmcouncil.org/fibo/ontology} for more information.} For instance, given the set of concepts “Bonds”, “Unclassified”, “Share”, “Loan”, the most relevant hypernym of “European depositary receipt” is “Share”. Figure \ref{fig:hypernymy} illustrates hypernym examples based on small parts of the FIBO ontology.

In this paper, we present our solution to the \finsim-3 task. Our system starts by augmenting all given terms with their Investopedia definitions. Then, we employ a Logistic Regression classifier over financial word embeddings and a mix of hand-crafted and distance-based features. Moreover, for the first time in this task, we explore various replacement methods for out-of-vocabulary (OOV) terms, leading to improved performance in our experiments. Our best-performing submission ranked 4th on the task’s leaderboard.

In what follows, Section \ref{sec:related_work} gives a brief overview of the related work in the context of the previous \finsim tasks. Section \ref{sec:data} presents the data that are provided in this task. Then, Section \ref{sec:system} presents our solution to the task, while Section \ref{sec:experiment} provides empirical evaluation results. Finally, Section \ref{sec:conlusions} summarizes the paper and presents future directions.

\section{Related work}
\label{sec:related_work}

\vspace*{1.5mm}
\noindent\textbf{Hypernym modeling:} Unsupervised hypernym modeling and classification mainly rely on measures assuming a distributional inclusion. This means that if a term $c$ is semantically narrower than term $p$, then a number of distributional features of $c$ should also be included in the feature vector of $p$ \cite{Lenci2012}. Similarly, the work in \cite{Santus2014} is based on the distributional informativeness hypothesis, which assumes that hypernyms tend to be less informative than hyponyms. Such distributional approaches rely on vector semantics and represent words as vectors.

Supervised methods are mainly based on word embeddings to represent words as low dimensional vectors in a latent space. Hypernym/hyponym pairs are encoded as combinations of two-word vectors, and hypernym relation classification is usually performed by training a classifier given the latter combinations of vectors as input \cite{Baroni2012,Roller2014,Weeds2014}. Other approaches rely on pre-extracted taxonomic relation data to create word embeddings that are later used as input to a Support Vector Machine (SVM) to learn the hypernym relation \cite{Tuan2016,Yu2015}.

In the context of \finsim-3, we approach the task as a classification problem, aiming to classify input terms to their correct hypernym. We follow a supervised distributional approach without explicitly modeling hypernym relations or other ontological relations.

\vspace*{1.5mm}
\noindent\textbf{\finsim:} The first task to propose hypernym categorization in the financial domain was \finsim-1 \cite{finsim-1-proceedings}, having a total tagset of 8 FIBO classes/hypernyms.
The 1st-year winner system \cite{finsim-1-winners-iitk} combined rules and a Naive Bayes classifier over word2vec embeddings \cite{word-embeddings}, overperforming BERT \cite{bert} embeddings.
The runner-up system \cite{finsim-1-runners-up} augmented all terms with their Investopedia definition and used a linear SVM over some hand-crafted and bi-gram TF-IDF features.

One year later, \finsim-2 \cite{finsim-2-proceedings} held place, expanding the tagset to 10 financial hypernyms. The \finsim-2 winners \cite{finsim-2-winners-polyu} used a Logistic Regression classifier over word embeddings, semantic and string similarities, along with BERT-derived masking probabilities to classify each term to a hypernym. The second-place system \cite{finsim-2-runners-up-goat} also used a Logistic Regression classifier over fine-tuned word embeddings derived from various financial text corpora.

Our system augments all terms with their Investopedia definitions, following \cite{finsim-1-runners-up}, and incorporates hand-crafted and distance-based features. However, in contrast to previous works, we also experiment with different OOV replacement methods for unknown terms to deal with the many words in the dataset that are not contained in the vocabulary of the pre-trained word embeddings. 

\section{Data}
\label{sec:data}
In this section, we briefly present the data provided by the task organizers, which include the training dataset, the FIBO ontology, the prospectus corpus, and the word embeddings.

\vspace*{1.5mm}
\noindent\textbf{Dataset:} The dataset consists of one-word or multi-word concepts from the financial domain and their labels. It is separated into training and test sets, with the former being released a month before the test set. 

The training set comprises 1050 examples with their corresponding labels. The unique labels are 17 in total, and their frequencies in the training data are provided in Table~\ref{tab:class_dist} in descending order. The most frequent label is \emph{Equity Index}, which appears in 27\% of the training examples, while the rarest labels are \emph{Forward} and \emph{Securities restrictions}. Both of them appear less than 10 times in the training data.

\begin{table}
\centering
\caption{Class distribution in the training set.}
\label{tab:class_dist}
  \begin{tabular}{lr} 
    \toprule
    Class & Count\\
    \midrule
    Equity Index & 286\\
    Regulatory Agency & 205\\
    Credit Index & 129\\
    Central Securities Depository & 107\\
    Debt pricing and yields & 58\\
    Bonds & 55\\
    Swap & 36\\
    Stock Corporation & 25\\
    Option & 24\\
    Funds & 22\\
    Future & 19\\
    Credit Events & 18\\
    Stocks & 17\\
    MMIs & 17\\
    Parametric schedules & 15\\
    Forward & 9\\
    Securities restrictions & 8\\
\bottomrule
  \end{tabular}
\end{table}

\vspace*{1.5mm}
\noindent\textbf{FIBO Ontology:} The Financial Industry Business Ontology (FIBO) is a pioneering effort to formalize the semantics of the financial domain using a large number of ontologies. At the time of this writing, FIBO is still a work in progress. However, it already defines large sets of concepts that are of interest in financial business applications and how these concepts relate to one another.

For each concept, FIBO provides additional textual information, including a definition, a generated description, labels, titles, and in many cases, a small abstract. We combine the FIBO textual information with the prospectus corpus (see below) for training custom embeddings. In particular, we traverse the ontologies and concepts related to the labels in the training set, and we fetch the corresponding textual information. This way, we augment the prospectus corpus with additional concept-specific documents from the ontology that include small snippets, descriptions, and definitions.

\vspace*{1.5mm}
\noindent\textbf{Prospectus corpus:} A corpus of documents is also provided in the English language for training word embeddings. The corpus has been compiled from various websites, comprising financial prospectuses, and it consists of approximately 14M tokens. Those files are given in pdf format. We have used this corpus and text parts of the FIBO ontology to train custom word embeddings. 

\vspace*{1.5mm}
\noindent\textbf{Word embeddings:} Finally, two sets of pre-trained word embeddings are also provided in the context of the task. Both sets are based on the Word2Vec model and trained in an internal financial corpus by the organizers, which comprises key investor information documents and financial prospectuses. The difference between the two word embeddings is the number of dimensions of the vectors and the vocabulary size. The first comprises 100-dimensional vectors and 17328 words, while the second has 300 dimensions and 34437 words.

\section{System description}
\label{sec:system}

\begin{figure}[h]
  \centering
      \includegraphics[width=0.85\linewidth]{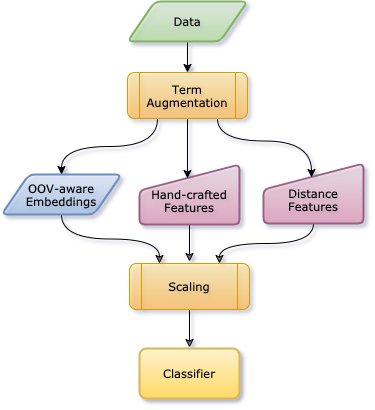}
  \caption{The pipeline of our best system.}
  \label{fig:system_arch}
\end{figure}

Our system extends baseline 2, which is a Logistic Regression classifier over the word2vec embeddings from the \finsim-3 organizers.
Figure \ref{fig:system_arch} illustrates the pipeline of our system. After augmenting all terms with their Investopedia definitions, we concatenate and scale the OOV-aware embeddings with the hand-crafted and distance features. We then use a Logistic Regression classifier over these features.

In the following subsections, we present how we worked towards building our pipeline.

\subsection{Feature Engineering}\label{subsec:feature_engineering}
First, we created a set of hand-crafted features that are indicative of specific classes. In order to gain insights on terms that could be indicative of each class, we did a preliminary error analysis using the provided baseline 2 system as a predictor. We devised 7 simple boolean hand-crafted features that denoted the existence of a specific string in the term. These strings were common in miss-classified terms of the baseline model while also being highly indicative of the true class of the term. For example, the occurrence of the string ``\textit{Inc.}'' was very common in errors of the baseline, while at the same time, almost all terms having this string belonged to the \textit{Stock Corporation} class. 

Moreover, specific classes such as \textit{Credit Index} have a lot of upper-case letters or unusual patterns. To capture these intricacies, we have also added 3 features that correspond to the number of characters in the term, the number of upper-case letters, and the ratio of upper-case to lower-case letters. We ended up with 10 such hand-crafted features.

 Furthermore, in order to represent the latent space distance between the terms and the classes, we calculated the cosine distance between the term's embedding and each class' embedding, adding 17 features in total.\footnote{The vector representation for multi-token terms/classes is the sum of each token's vector representation. We also tried the centroids of the embeddings which reduced the performance.} Last but not least, we calculated the Levenshtein distances between the term and each class label, adding another 17 features.
 After concatenating all of the features, we scale them to a uniform range of [-1, 1].
 
\subsection{Out-of-Vocabulary Words}\label{sub:oov_words}
Inside the 1050 training terms, we found at least 214 words that were out of vocabulary, using the 300-dimensional organizers' embeddings.
Such words were mainly instances of credit indexes and regulatory agencies.
Since terms contain a single word or a small number of words, we identified the need to deal with \oov occurrences instead of using zero embeddings. For example, the word "asiacorporate" and "t-bill" are \oov and they are represented by zero embeddings. Ideally, we would like to match them correctly to the most similar in-vocabulary words ``corporate" and ``treasury-bill" in order to retrieve the best vector representations available to help the classification process.

First, we tried replacing each \oov word with its closest in-vocabulary word in terms of Levenshtein distance, which helped performance.
Then, we replaced this relatively simple mechanism by utilizing the Magnitude toolkit \cite{patel-etal-2018-magnitude}. Magnitude works as a wrapper for already-trained word2vec models. It allows advanced OOV lookup as it combines different character n-grams, string similarities, and morphology-aware matching.

\subsection{In-domain representations}\label{sub:indomain_representations}
We use the 300-dimensional embeddings provided by the organizers for the word vector representations since they perform better in the development set than the 100-dimensional ones.

We also experimented with a wide variety of other domain-specific word representations.
First, we trained our financial word2vec embeddings (d=200) on the prospectus corpus and the FIBO ontology provided. We extracted the text from the prospectuses PDF files using the provided pdf-to-text toolkit\footnote{Consult \url{https://poppler.freedesktop.org/} for more information.}. 
However, they proved to be less beneficial than the provided embeddings.

Furthermore, we extracted the embedding layer of \finbert~\cite{yang2020finbert}, a financial BERT, and converted it to word2vec format.
To our surprise, the provided embeddings outperformed the \finbert embeddings. The provided embeddings also outperformed our custom word2vec and BERT embeddings, trained on financial documents from the Securities \& Exchange Commission.

\subsection{Augmentation with Investopedia Terms}\label{sub:investopedia_augmentation}
During the later stages of our experimentation, we noticed that many of the misclassifications were attributed to acronyms (e.g. "\textit{FRN}") or common words being present ("\textit{Long call}/\textit{put}"). In order to alleviate this problem and provide more context for all terms, we utilized Investopedia\footnote{\url{https://www.investopedia.com/}} as a dictionary of definitions for each term. To do this, we built a scrapper that pinpoints the closest match of a given term that has a definition in the terms dictionary of the website. The scrapper first tries to find exact matches of the query term in the dictionary of the website. If this fails, we utilize the search functionality of the site to identify the closest matching term. Having found a corresponding match (exact or approximate), we fetch the definition of the matched term and keep only the first sentence of the definition, as it usually is in the format: "\textit{[Term] is a ..}", where \textit{[Term]} denotes the matched term for the given query term.

This process is followed both for augmenting the initial training data and the final test terms, as it can be easily incorporated at inference time. Following the above process, approximately $\ 70\%$ of both the train and test terms were augmented. In case the augmentation process did not retrieve any definition for a given term (this was common for \textit{Credit Indexes} for example), the term was left as is.

\subsection{Classifier tuning}\label{sub:classifier_tuning}
Apart from Logistic Regression, we also evaluated a battery of different classifiers implemented in the scikit-learn library, like the Naive Bayes Classifier, Decision Trees, linear SVMs, Multi-Layer Perceptron, XGBoost, and RUSBoost \cite{RusBoost}, without any improvement in performance.
Indicatively, classifiers based on trees (Decision Trees, XGBoost, RUSBoost) scored the lowest, possibly due to the extensive feature space.

Thus, we chose to continue with the simple -yet powerful- Logistic Regression classifier, where we tuned the regularization strength hyperparameter \textsc{C}. We defined a search space of $\{0.001, 0.01, 0.1, 1.0, 10.0, 100.0\}$. We found that C=$0.1$ is the best option in terms of mean rank and accuracy based on a stratified 5-fold cross-validation setting.

\section{Experimentation}
\label{sec:experiment}

\subsection{Metrics}\label{sub:metrics}
We evaluate our performance using accuracy, mean rank and macro-average F1 score, as shown in equations \ref{eq:acc}, \ref{eq:rank} and \ref{eq:f1}.

\begin{equation}\label{eq:acc}
    Accuracy = \frac{TP+TN}{TP+TN+FP+FN}
\end{equation} 

\begin{equation}\label{eq:rank}
    Mean Rank = \frac{1}{n} * \sum_{i=1}^{n}rank_i
\end{equation}

\begin{equation}\label{eq:f1}
    Macro F1 = \frac{1}{n} * \sum_{i=0}^{n} {F1}_i
\end{equation}

In the context of the shared task, apart from accuracy, we also had to generate all labels in ranked order and measure the mean rank. For each term $x_i$ with a label $y_i$ from the n samples in the test set, the expected prediction is a top-3 list of labels ranked from most to least likely to be equal to the ground truth. In equation \ref{eq:rank}, $rank_i$
is the rank of the correct label in the top-3 prediction list. If the ground truth does not appear in the top-3 then $rank_i$ is equal to 4.


\subsection{Results}\label{sub:results}
Table \ref{tab:experiments} presents the experimental results of our system's variations. We used the Logistic Regression classifier in a stratified 5-fold cross-validation setting. Since the  training set is small and imbalanced we selected that setting in order to ensure that all classes will be represented in each fold. We provide empirical results using the best hyperparameters after tuning (see Subsection \ref{sub:classifier_tuning}). In particular, the following variations were evaluated:

\begin{itemize}
    \item \textbf{BL}: The baseline model. Logistic Regression classifier with the given input embeddings as features.
    \item \textbf{BL.HF}: BL with additional hand-crafted features (Subsection \ref{subsec:feature_engineering}).
    \item \textbf{BL.HF.OOV\textsubscript{l}}: BL.HF plus the Levenhstein-based OOV words handling  (Subsection \ref{sub:oov_words}).
    \item \textbf{BL.HF.OOV\textsubscript{l}.D}: BL.HF.OOV\textsubscript{l} plus additional features based on the cosine distance between term and class embeddings (Subsection \ref{subsec:feature_engineering}).
    \item \textbf{BL.HF.OOV\textsubscript{l}.D\textsuperscript{2}}: BL.HF.OOV\textsubscript{l} plus 
    the character distance between terms and classes (Subsection \ref{subsec:feature_engineering}).  
    \item \textbf{BL.HF.OOV\textsubscript{m}.D\textsuperscript{2}}: BL.HF plus the Magnitude-based OOV words handling 
    (Subsection \ref{sub:oov_words}). This variation constitutes the first submission of our system (\dicoe1) to the shared task.
    \item \textbf{BL.HF.OOV\textsubscript{m}.D\textsuperscript{2}.+}: This is BL.HF.OOV\textsubscript{m}.D\textsuperscript{2} using Investopedia-based augmented terms (Subsection \ref{sub:investopedia_augmentation}). This variation constitutes the second submission of our system (\dicoe2) to the shared task.
\end{itemize}

\begin{table}
\centering

\caption{
Experimental results based on stratified 5-fold cross validation. Results are shown using a tuned Logistic Regression Classifier (C=0.1).
}
\label{tab:experiments}
  \begin{tabular}{lrrr} 
    \toprule
    Model & Mean Rank & Accuracy & Macro F1\\
    \midrule
    BL & 1.196 & 87.6 & 80.0\\
    BL.HF & 1.166 & 90.0 & 82.0\\
    BL.HF.OOV\textsubscript{l} & 1.156 & 90.5 & 83.0\\
    BL.HF.OOV\textsubscript{l}.D & 1.148 & 90.7 & 84.0\\
    BL.HF.OOV\textsubscript{l}.D\textsuperscript{2} & 1.147 & 90.8 & 83.8\\
    BL.HF.OOV\textsubscript{m}.D\textsuperscript{2} & 1.144 & 91.2 & 84.1\\
     \textbf{BL.HF.OOV\textsubscript{m}.D\textsuperscript{2}.+} & \textbf{1.132} & \textbf{91.5} & \textbf{85.0}\\
     
\bottomrule
  \end{tabular}
\end{table}

Table~\ref{tab:experiments} shows that incorporating the hand-crafted features improves the baseline by 2.4\% in terms of accuracy. The mean rank also improves from 1.196 to 1.166, suggesting that simple substring binary features may indicate the specific class of each term. Then, we first leverage the Levenshtein distance to replace each OOV word found in the terms with its closest in-vocabulary word. This boosts accuracy by 0.5\%, while the mean rank is reduced to 1.156. Thus, handling OOV words with replacements allows us to retrieve better vector representations than zero embeddings. Our next improvement combines the Levenshtein character distance between the term's words and the class labels, as well as the cosine distance between their representations in the latent space, improving accuracy to 90.8\% and mean rank to 1.147. 

Next, we combine all of the above and replace the simple Levenshtein OOV mechanism with the Magnitude toolkit~\cite{patel-etal-2018-magnitude}. This advanced OOV lookup method scored 91.2\% in terms of accuracy and 1.144 in terms of mean rank and represents our first submission to the \finsim-3 shared task (see BL.HF.OOV\textsubscript{l}.D\textsuperscript{2} in Table~\ref{tab:experiments}). 

Our final system extends the first submission by augmenting the financial terms with their Investopedia definitions (Section~\ref{sub:investopedia_augmentation}) in order to provide more context for the classification. This was our best system and scored a 1.132 mean rank, 91.5\% accuracy, and 85.0\% Macro F1 Score in the 5-fold cross validation (see BL.HF.OOV\textsubscript{m}.D\textsuperscript{2}+ in Table~\ref{tab:experiments}).

\section{Conclusions and Future Work}\label{sec:conlusions}
We presented \dicoe team's submissions to \finsim-3. Our  Investopedia-augmented system ranked 4th on the leaderboard. We leveraged hand-crafted and distance-based features which led to significant improvements over the baseline. To our surprise, external and modern financial word representations, such as \finbert, did not contribute positively to the results. Moreover, for the first time in this shared task, we introduced the application of OOV word replacement methods. Using OOV replacements, we can successfully retrieve correct vector representations for unknown tokens that share the same morphology with in-vocabulary words.

In future work, we plan to investigate other ways of augmenting terms with their definitions and broad context, as well as creating new external financial resources for experimentation. An additional future direction is the direct modeling of the hypernym relation using pairs of tokens and labels to explicitly learn that type of relation.



\bibliographystyle{named}
\bibliography{ijcai21}

\end{document}